\documentclass[letterpaper, 10 pt, conference]{ieeeconf}  

\IEEEoverridecommandlockouts                              

\usepackage{graphicx} 
\usepackage[utf8]{inputenc}
\usepackage{amsmath,amsfonts,booktabs,cite} 
\usepackage{gensymb} 
\usepackage{siunitx,textcomp} 
\usepackage[hidelinks]{hyperref} 
\usepackage{cleveref} 
\usepackage{subcaption}
\usepackage{tabularx}
\usepackage{makecell}
\usepackage{pbox}
\let\labelindent\relax
\usepackage[inline]{enumitem} 
\usepackage[nolist]{acronym} 
\usepackage{multirow}
\usepackage[ruled,vlined]{algorithm2e}
\usepackage{bm}

\usepackage[font=small]{caption}

\def\figref#1{Fig.~\ref{#1}}

\def\eqref#1{Eq.~(\ref{#1})}

\newcommand\etal{~\emph{et al. }}
\newlength{\twosubht}
\newsavebox{\twosubbox}

\crefname{algocf}{alg.}{algs.}
\Crefname{algocf}{Algorithm}{Algorithms}

\title{\LARGE \bf
Graph-based View Motion Planning for Fruit Detection
}

\author{Tobias Zaenker \and  Julius R\"{u}ckin \and Rohit Menon \and Marija Popovi\'{c} \and Maren Bennewitz
\thanks{This work has partially been funded by the Deutsche Forschungsgemeinschaft (DFG, German Research Foundation) under Germany’s Excellence Strategy –  EXC-2070 -- 390732324 -- Phenorob and under the grant number BE 4420/4-1 (FOR 5351: KI-FOR Automation and Artificial Intelligence for Monitoring and Decision Making in Horticultural Crops, AID4Crops). Tobias Zaenker, Rohit Menon, and Maren Bennewitz are with the Humanoid Robots Lab, Julius R\"{u}ckin and Marija Popovi\'{c} are with the Institute of Geodesy and Geoinformation, University of Bonn, Germany. Maren Bennewitz is additionally with the Lamarr Institute for Machine Learning and Artificial Intelligence, Germany.}}

\begin{document}

\maketitle
\thispagestyle{empty} 
\pagestyle{empty}

\begin{abstract} 


Crop monitoring is crucial for maximizing agricultural productivity and efficiency.
However, monitoring large and complex structures such as sweet pepper plants presents significant challenges, especially due to frequent occlusions of the fruits.
Traditional next-best view planning can lead to unstructured and inefficient coverage of the crops.
To address this, we propose a novel view motion planner that builds a graph network of viable view poses and trajectories between nearby poses, thereby considering robot motion constraints. The planner searches the graphs for view sequences with the highest accumulated information gain, allowing for efficient pepper plant monitoring while minimizing occlusions. The generated view poses aim at both sufficiently covering already detected and discovering new fruits. The graph and the corresponding best view pose sequence are computed with a limited horizon and are adaptively updated in fixed time intervals as the system gathers new information. 
We demonstrate the effectiveness of our approach through simulated and real-world experiments using a robotic arm equipped with an RGB-D camera and mounted on a trolley.
As the experimental results show, our planner produces view pose sequences to systematically cover the crops and leads to increased fruit coverage when given a limited time in comparison to a state-of-the-art single next-best view planner.

\end{abstract}

\section{Introduction}
\label{sec:intro}


Advances in robotic automation enable highly flexible and low-cost operation and intervention, such as in precision agriculture~\cite{rodriguez2021mapping} and human-aware service robots~\cite{carlson2008human}.
Endowing these robots with recent progress in deep learning for automated semantic interpretation of scene observations~\cite{long2015fully, badrinarayanan2017segnet} is a promising avenue towards deployment in challenging applications.
To bridge the gap between environment perception and intervention, a key robotic skill is to acquire informative observations about the environment to solve complex tasks, such as fruit harvesting~\cite{zhou2022intelligent}.
Particularly, the robot has to plan where to measure next to maximize information about the environment while reasoning about incomplete or uncertain observations, e.g., due to occlusion of fruits by leaves.

This work examines the problem of view pose planning under partial observability in 3D environments using a robot arm equipped with an RGB-D camera in a restricted 3D workspace.
We aim at mapping initially unknown plant rows in a glasshouse with an arm mounted to a trolley maximizing the number of covered fruits given limited mission time.

Traditional approaches naively explore the unknown scene either randomly or cover it in a homogeneous non-targeted fashion~\cite{galceran2013survey}.
Such \textit{non-adaptive} methods fail to focus on informative regions as they are discovered online and cannot reason about partially observed fruits or adapt to the robot arm's motion capabilities.
\textit{Adaptive} approaches have achieved promising results optimizing a single next-best view pose given the current environment map~\cite{zaenker2020ecmr, zaenker2020viewpoint, menon2022viewpoint}.
However, these approaches rely on myopic greedy optimization of single next-best view poses, which is prone to result in suboptimal paths under partial observability.
Furthermore, none of the previous works integrate the robot arm's motion planning into the view pose optimization procedure.
Thus, these methods tend to plan paths that are inefficient to execute in restricted workspaces, such as in glasshouses.

\begin{figure}[!t]
	\centering
	\includegraphics[width=\linewidth]{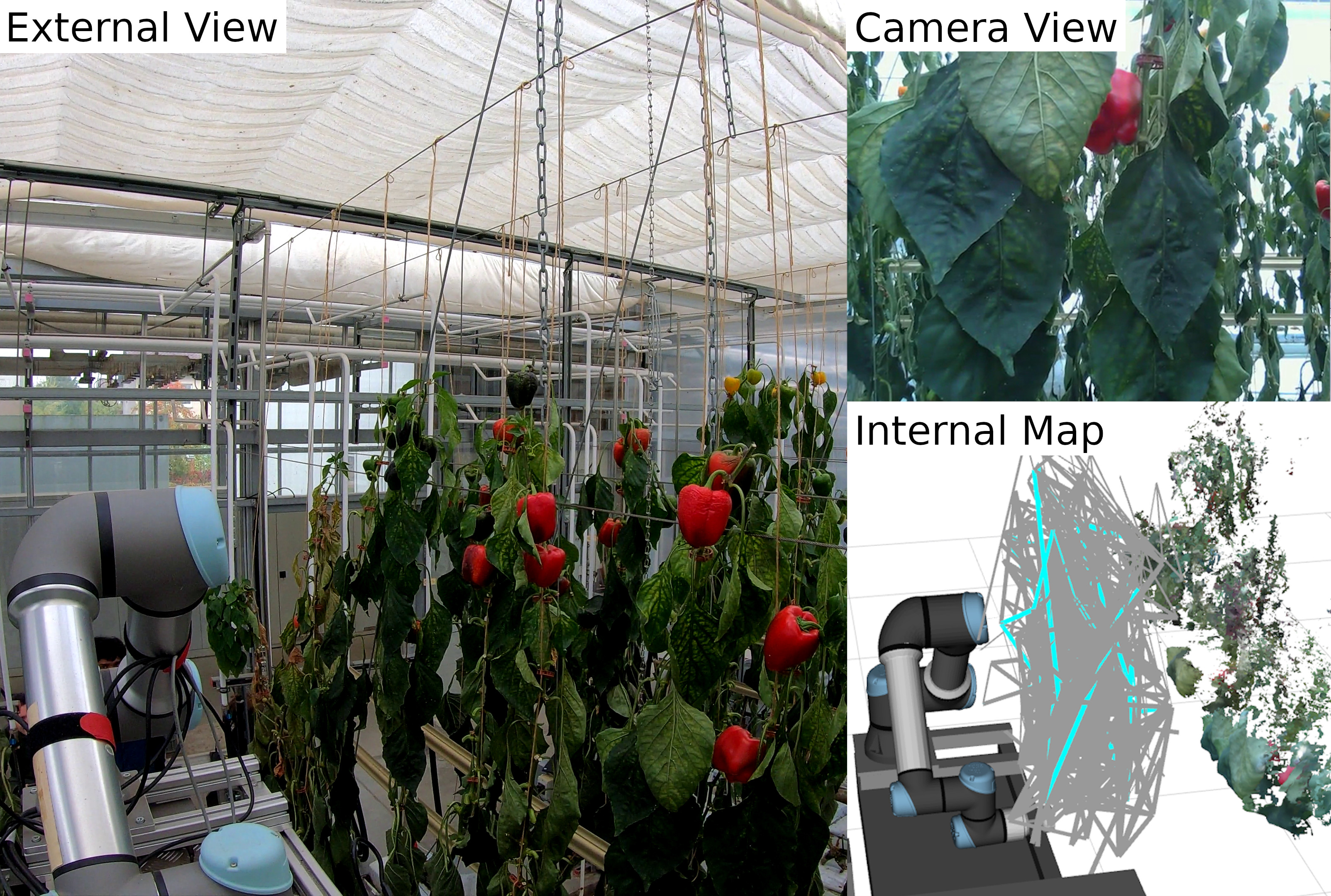}	
	\caption{Our system deployed in the glasshouse.
		On the left, the system setup is captured by an external camera.
		The top right shows the current view of the arm camera.
		On the bottom right, the internal map of the environment can be seen, as well as the view pose graph generated by our planning algorithm.
		Connected view poses are visualized with gray lines, while the light blue line shows the currently predicted best view pose sequence to cover the current section.
    }
	\label{fig:coverfig}
\end{figure}

Therefore, we present a graph-based planner that generates information-rich paths, i.e., sequences of view poses, to maximize the number of newly discovered fruit clusters while considering the arm's motion constraints.
Our approach replans the view pose sequence online as new observations arrive to \textit{adaptively} target fruits not yet sensed or not yet sufficiently covered.
To achieve this, we propose a new planning procedure combining view pose information estimation and the robot arm's motion planning to ensure both information-rich and efficiently executable paths.
The key benefit of our approach is its high data collection efficiency in restricted robot arm workspaces.

In sum, (i) we show that our view pose planner increases the number of detected fruit clusters in restricted robot arm workspaces with a limited mission time budget compared to a state-of-the-art next-best view planner; (ii) we verify that our system can be readily deployed in real-world scenarios accounting for the robot arm's motion constraints where the previous method is prone to fail to plan collision-free paths; (iii) to support reproducibility, we open-source our modular and easy-to-integrate ROS implementation of the complete system: \url{https://github.com/Eruvae/view_motion_planner}.

\section{Related Work}
\label{sec:related}


View pose planning has been widely studied in robotic applications, e.g., precision agriculture~\cite{zaenker2020viewpoint,sukkar_multi-robot_2019,menon2022viewpoint}, environmental exploration and monitoring~\cite{choudhury2020adaptive,dang2019graph,rodriguez2021mapping}, and household robotics~\cite{monica2018contour}.
Traditional \textit{non-adaptive} methods, e.g., coverage path planning~\cite{galceran2013survey}, compute a fixed path uniformly covering the environment not focusing on regions of interest. These methods assume a known static environment rendering them impractical for autonomous exploration in initially unknown environments.

In contrast, \textit{adaptive} methods, such as our proposed approach, support online replanning based on incoming sensor data.
We distinguish between (i) mapless methods, which plan based only on sensor data, and (ii) map-based methods, which consider a global environment representation subsequently leveraged for planning.
In the former category, Lehnert\etal\cite{lehnert2016sweet} proposed an approach for guiding a camera array in a direction to improve target visibility in occluded environments.
Zhan\etal\cite{Zhan2022} introduced an object-centric next-best view learning framework exploiting implicit neural representations to improve 3D scene reconstruction.
Lauri\etal\cite{Lauri2020} developed a greedy planner minimizing the predicted overlap between multiple depth cameras.
In general, mapless methods improve scalability for online planning in larger environments. However, as they rely on locally greedy next-best view pose optimization instead of non-myopic planning of view pose sequences, they are susceptible to local minima in geometrically complex environments.



For map-based planning, we focus on methods leveraging volumetric maps~\cite{oleynikova2017voxblox,Delmerico2018} to estimate expected information gain as in our approach.
Various works study myopic map-based next-best view planning, i.e., greedily planning only one step using sampling~\cite{lehnert2016sweet,zaenker2020viewpoint,menon2022viewpoint} or heuristic motion primitives~\cite{monica2018contour,palazzolo2018effective} to generate candidate view poses.
In a similar problem setup to ours, Zaenker\etal\cite{zaenker2020ecmr} combined global map-based sampling and local image-based occlusion avoidance to greedily find the next-best views maximizing coverage.
For manipulator planning, Wang\etal\cite{wang_autonomous_2019} proposed combining a hand-crafted map entropy metric with an information gain learned directly from depth images.
However, such myopic approaches share the disadvantages of mapless methods since they rely on a one-step lookahead.

Recent works show that non-myopic graph-based planning methods yield promising performance across various tasks and domains~\cite{amiri2022reasoning, choudhury2020adaptive, dang2019graph}. 
Amiri\etal\cite{amiri2022reasoning} proposed a graph-based planner for task completion under partial observability in indoor environments. 
Choudhury\etal\cite{choudhury2020adaptive} introduced an adaptive informative path planning method leveraging partially observable Monte Carlo planning~\cite{silver2010monte} over graphs encoding a UAV's workspace. 
Dang\etal\cite{dang2019graph} developed a system for autonomous graph-based exploration in subterranean environments combining locally bounded rapidly exploring random graphs and global planning to foster exploration of unknown space.
Despite the widespread applications of graph-based non-myopic planning, to the best of our knowledge, our work is the first to consider graph-based planning in view pose optimization for targeted fruit monitoring. We propose a two-step planning approach. First, we build a graph of candidate view poses. Then, we apply a non-myopic best-first graph search to maximize detected fruit clusters. In contrast to previous works in other domains, our graph-based planner combines both information gain optimization and motion planning to guarantee dynamically feasible and efficiently reachable paths.

A large body of research tackles view pose planning for crop or fruit monitoring to support selective harvesting and yield estimation tasks.
Sukkar\etal\cite{sukkar_multi-robot_2019} proposed an approach for observing apples with multiple robot arms using Monte Carlo tree search. Similar to us, they plan in a non-myopic fashion and evaluate candidate view poses based on the predicted visibility of the fruit. A key difference is that their approach does not utilize the global map to sample targeted view poses, whereas we adaptively guide the sampling procedure towards regions of interest. 
Van Essen\etal\cite{van2022dynamic} use an attention-driven approach to determine the next-best views for 3D plant reconstruction. However, their problem setup is different from ours since they consider a plant-centric environment while we consider a moving trolley in a cluttered glasshouse.
Most similar to our work, Zaenker\etal\cite{zaenker2020viewpoint} introduced an adaptive sampling-based planner for sweet pepper reconstruction. We build upon their approach by incorporating graph-based non-myopic planning catering to the robot's motion constraints. This allows us to achieve more informative and efficiently executable view pose sequences in restricted workspaces.

\section{System Overview}
\label{sec:approach}

\begin{figure}[!t] 	
    \centering 	
    \includegraphics[width=\linewidth]{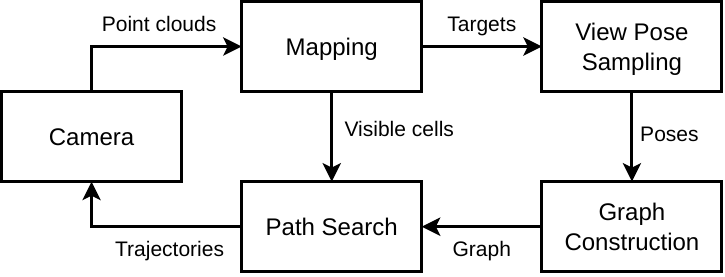} 	
    \caption{Overview of our system. We use OctoMap, a probabilistic 3D map, to fuse point clouds updating the map state of unknown, occupied, free space, and regions of interest (fruits). Based on the map state, we sample potentially informative target positions subsequently used in our novel view pose planner. The planner constructs a graph of efficiently reachable view poses around target positions and performs a best-first graph search in parallel to find informative and quick-to-execute view pose sequences.} 
    \label{fig:system_overview}
\end{figure}

We present a novel graph-based view pose planning approach maximizing the information gained over regions of interest in initially unknown 3D environments. We consider a robot arm mounted to a trolley moving along plant rows in a glasshouse. Our goal is to maximize fruit coverage with limited mission time in constrained workspaces.
\figref{fig:system_overview} overviews our proposed view pose planning system. Similar to our previous work~\cite{zaenker2020viewpoint}, we utilize OctoMap~\cite{hornung13auro} to fuse point cloud observations from the camera into a probabilistic 3D map capturing unknown, occupied, free space, and regions of interest, i.e., fruit clusters. Based on the current map state, we sample target positions with potentially high information value, e.g., fruits or frontiers of known space. Our new view motion planner samples efficiently reachable view pose candidates around these target positions accounting for the robot arm's motion constraints. This way, we achieve fast path execution in restricted workspaces, such as glasshouses. Reachable nearby view poses are connected to a view pose candidate graph. A key feature of our approach is the non-myopic graph-based best-first path search.
The planned path is executed for a fixed time interval before we replan a new path to \textit{adaptively} target fruit clusters as they are discovered. To improve computation time, we execute graph construction and path search in parallel. \Cref{S:planning_approach} overviews our path planning approach.

\section{Planning Approach} \label{S:planning_approach}

Our key idea is to disentangle path planning into two highly parallelizable procedures.
First, we construct a view pose candidate graph.  We sample view poses facing potentially informative target positions, e.g., detected fruits, given the current map state. 
To achieve high path execution speed in restricted workspaces, we account for the robot arm's motion constraints while sampling view pose candidates. 
Second, we plan view pose sequences utilizing a best-first path search over the view pose candidate graph to maximize fruit coverage. To ensure informative and quick-to-execute paths, our new planning objective combines view pose utility estimation and motion planning. 
To adaptively target newly discovered fruits, we replan paths after a fixed time has elapsed.
This way, our planning approach enables efficiently executable paths in workspace-constrained glasshouses and maximizes monitored fruit clusters under limited mission time. The following subsections describe our graph construction and path search.

\subsection{Sampling-Based Graph Construction} \label{S:graph_building}


An important basis for our view pose planning approach is a constantly evolving graph generated by sampling view pose candidates facing potentially informative target positions, e.g., detected fruits. The extraction of target positions is updated based on the current map state, i.e., as new point cloud observations arrive. We adapt OctoMap~\cite{hornung13auro} to our fruit mapping task and sequentially fuse point cloud observations and their associated fruit detections into a global probabilistic 3D map maintaining our current belief about unknown, occupied, free space, and regions of interest, i.e., fruit clusters. The following subsections describe the sampling-based target position generation, sampling of view pose candidates around a given target position, and the graph construction based on the view pose candidates.

\begin{figure}[!t] 	
    \centering 	
    \includegraphics[width=\linewidth]{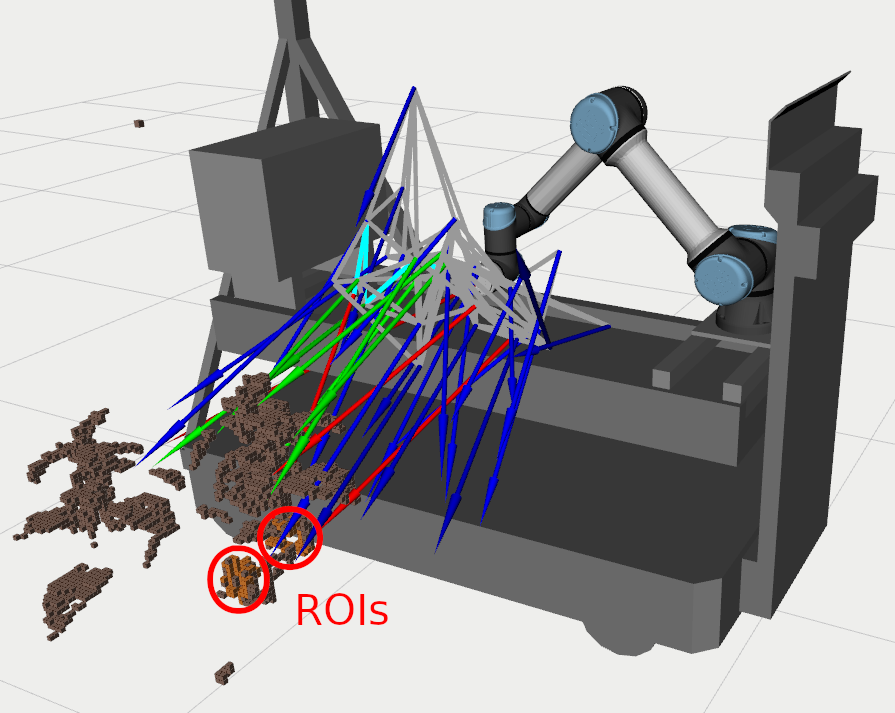} 	
    \caption{View pose candidate graph illustrating view poses facing three different target position types. First, we sample view poses facing regions of interest (ROIs), i.e., fruits (red arrows). Second, we sample view poses targeting occupied space, i.e., plants with potentially new fruits (green arrows). Third, we sample view poses in direction of frontiers of unknown space to foster exploration (blue arrows). The gray edges connect nearest-neighbor view poses. Light blue depicts the most informative planned path.}
    \label{fig:graph_vp_types}
\end{figure}

\subsubsection{Target Position Sampling}
We sample three different types of target positions with user-defined probabilities specifying the trade-off between exploring the scene and exploiting detected regions of interest.
First, we sample target positions at the frontiers between regions of interest and unknown space to complete potentially not yet monitored fruit parts.
Second, we sample target positions at the frontiers between occupied and unknown space to foster exploration of plants potentially carrying new fruits.
Third, we sample target positions at the frontiers between free and unknown space to foster exploration of the whole scene. 
We re-sample the set of target positions after each new point cloud observation is fused into the map. 
Then, we repeatedly choose a target position uniformly at random from all sampled targets.
Around the chosen target position, we sample a view pose candidate that is efficiently reachable by the robot arm.

\subsubsection{View Pose Candidate Sampling}


In our previous work \cite{zaenker2020viewpoint}, we sample view pose candidates from targets by casting rays in random directions within a user-defined range from the targets, and checking the rays for occlusions to obtain views with visibility of the target. This works well in open, relatively unconstrained setups, where the robot arm's workspace is large and sampled view poses can be reached with high probability.

However, in more restricted setups, there is a high likelihood of the sampled view being unreachable or lying outside of the arm's workspace. This case is most common in real-world glasshouses, like our setup where the robot arm is mounted on a trolley following plant rows. Therefore, we changed the approach to sample view poses at random from the robot arm's predefined workspace, looking towards a given target. Ray-casting is applied afterwards to check for potential occlusions and discard views where the target is not visible. Collision-free reachable view pose candidates are stored for subsequent graph construction.

\subsubsection{Graph Construction}
We utilize previously sampled view pose candidates to construct a graph subsequently used to search the view pose sequence maximizing fruit coverage. \figref{fig:graph_vp_types} exemplarily shows sampled view pose candidates for the three target position types connected to a nearest-neighbor pose graph. Each view pose candidate is inserted as a node into the graph together with its robot arm configuration.
Then, we insert edges from each node to its $k$ nearest-neighbor nodes with minimal robot arm joint angle configuration difference and point-wise collision-free arm trajectory.
For each collision-free trajectory from one view pose to another, we compute its execution time using MoveIt~\cite{moveit} and store it as additional edge information exploited during the path search to find efficiently executable view pose sequences.
Similarly, we add the camera pose as a node and connect it to its nearest-neighbor view poses. 

By construction, the graph consists of view poses that are efficiently reachable in glasshouses and provide potentially informative observations balancing exploration and fruit coverage. The following subsection introduces our path search over the view pose candidate graph.

\subsection{Graph-Based Path Search}
\label{sec:path_search}

\begin{algorithm}[t]
	\SetAlgoLined
        OctoMap state M\;
	View pose candidate graph G\;
	Priority queue P; \tcp{sorted by utility}
	P.insert(G.camNode())\;
    C(G.camNode()) = \{\}; \tcp{visible cell set}
	\While{not P.isEmpty()}
	{
		view\_\,pose = P.pop()\;
		\For{nb\_\,pose : G.neighbors(view\_\,pose)}
		{
			\If{nb\_\,pose.isExpanded()}
			{
                    continue\;
                }
                nb\_\,pose.addPredecessor(view\_\,pose)\;
                nb\_\,pose.setAsExpanded()\;
                visible\_\,cells = M.castRays(nb\_\,pose)\;
                C(nb\_\,pose) = C(view\_\,pose) + visible\_\,cells\; 
                nb\_\,pose.computeUtility(C)\;
                P.push(nb\_\,pose)\;
		}
	}
	\caption{Path search}
	\label{algo:path_search}
\end{algorithm} 

\Cref{algo:path_search} summarizes our approach to search a path in the view pose candidate graph introduced in \cref{S:graph_building}. 
\begin{figure}[b]
    \centering
    \includegraphics[width=0.98\columnwidth]{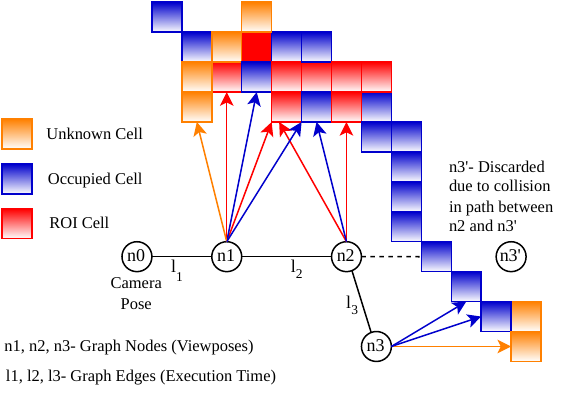}
    \caption{View motion graph with view poses as the nodes and execution times as the edges of the graph with an illustrative representation of the occupancy information.}
    \label{fig:vmp_graph}
\end{figure}
We start the path search procedure from the camera pose node.
We recursively expand nodes along the graph, similar to the classical depth-first search. At each expanded view pose node, we compute the visible unknown, free, occupied, and region of interest cells by casting rays into the current map state.
As we traverse the graph, map cells are visible from multiple view poses along the path $\psi_t = (n_1, \ldots, n_t)$ of length $t$, as illustrated in \figref{fig:vmp_graph}. Thus, in each node expansion step $t$, we insert the visible map cells into a set of unique map cells $C(n_t)$ visited while traversing the graph up to the current node $n_t$.
Based on the set of visible map cells $C(n_t)$, we estimate the node's utility $U(n_t)$ proposing a new planning objective combining information gain estimation and the path's execution time: 
\begin{equation}
	U(n_t) = \frac{C(n_t)_\mathit{unk}}{L(\psi_t)}\, \frac{(\mathit{ROI}(\psi_t) + 1)}{(t + 1)} \, ,
\end{equation}
where $C(n_t)_\mathit{unk}$ is the number of uniquely visible unknown cells while traversing $\psi_t$, $\mathit{ROI}(\psi_t)$ is the number of view poses in $\psi_t$ facing towards detected fruits and $L(\psi_t)$ is the path execution time summed along edges of $\psi_t$.
This way, we guide our planner towards regions of interest, i.e., fruit clusters, and at the same time foster exploration. Normalizing the information gain estimate by the path's execution time ensures efficient arm trajectories, especially in strongly restricted robot arm workspaces.  

The expanded node $n_t$ is inserted into a priority queue with priority $U(n_t)$.
In the next iteration, the node with the highest utility is extracted, and its not yet expanded neighbors are expanded in a best-first fashion.
For each expanded view pose, we store its predecessor with the highest utility up to this node. To find the to-be-executed view pose sequence maximizing the utility, we backtrack these stored predecessors starting at the view pose with the highest utility so far until the camera pose is reached.

The planned path is executed for a maximum look-ahead distance. 
During path execution, we fuse observations obtained at planned view poses into the map. Before executing the next view pose, we check for collisions based on the updated map state.
In case of collisions, we abort the path execution, remove the corresponding edge from the graph, and replan a new path based on the updated graph as illustrated in \figref{fig:vmp_graph}.
In case the reached camera pose differs from the planned view pose, we insert a collision-free edge from the camera pose to the planned view pose. If no collision-free edge can be inserted, we replan the path starting from the current camera pose.

\section{Experiments}
\label{sec:exp}

We compare our view motion planning (\textbf{VMP}) approach to our previously developed single next-best region of interest (ROI) view planner (\textbf{RVP})~\cite{zaenker2020viewpoint} in a simulated scenario to evaluate the generated view poses and the corresponding fruit discovery capabilities. Furthermore, we conducted experiments in a commercial glasshouse environment to qualitatively compare the resulting paths.

\subsection{Experimental Setup} \label{s:exp_setup}

\begin{figure}
	\centering
	\sbox\twosubbox{
		\resizebox{\dimexpr.96\linewidth}{!}{
			\includegraphics[height=2.6cm]{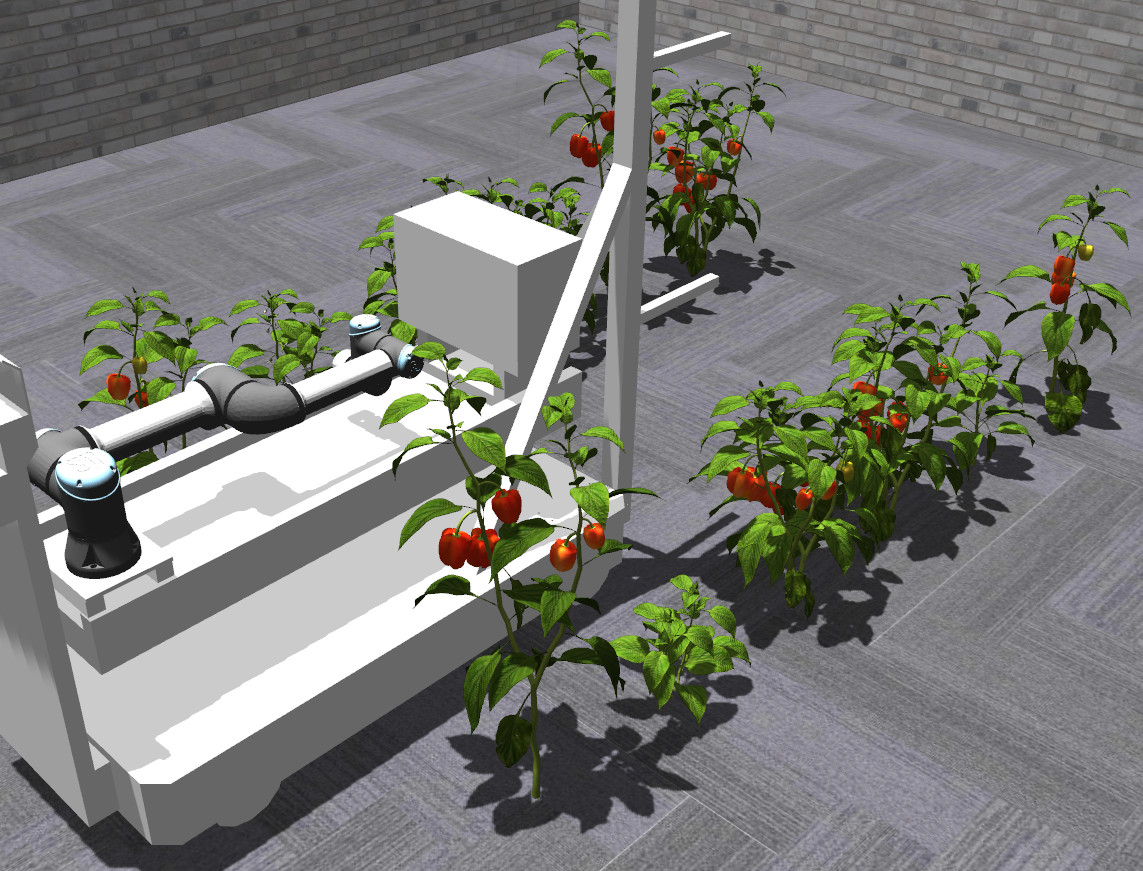}
			\includegraphics[height=2.6cm]{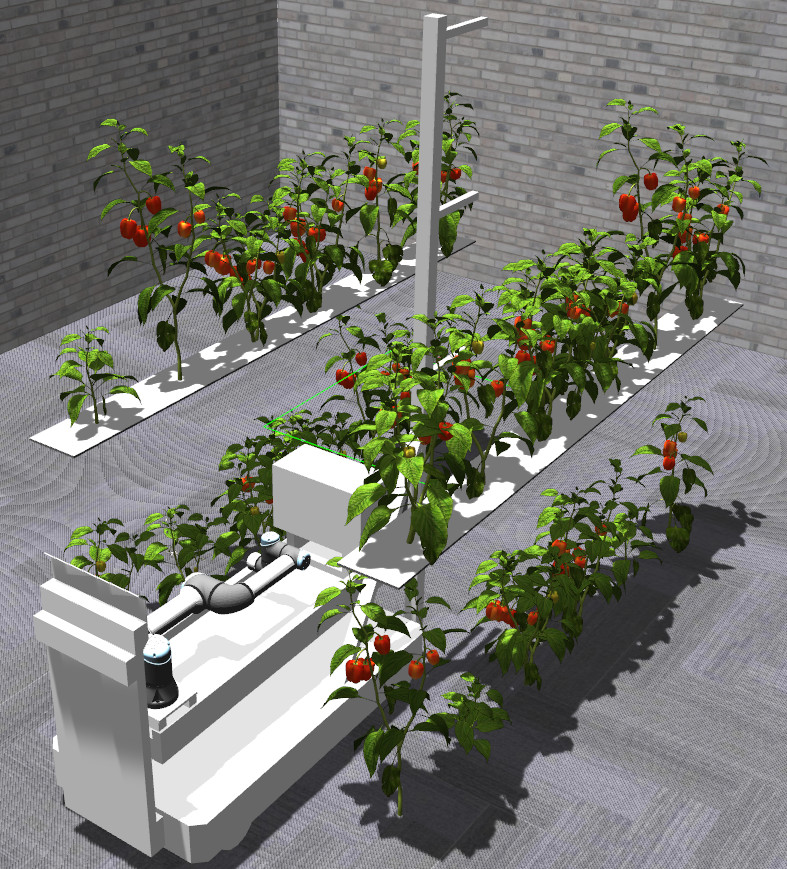}
		}
	}
	\setlength{\twosubht}{\ht\twosubbox}
	\centering
	\subcaptionbox{Scenario 1\label{fig:simulated_env_w1}}{
		\includegraphics[height=\twosubht]{images/world22}
	}\quad
	\subcaptionbox{Scenario 2\label{fig:simulated_env_w2}}{
		\includegraphics[height=\twosubht]{images/world23}
	}
	\caption{Glasshouse simulation scenarios. \textit{Left}: 12 plants arranged in two rows with a total of 47 sweet pepper fruits. \textit{Right}: 24 plants over two levels with two rows each and 94 sweet pepper fruits in total. A robot arm equipped with an RGB-D camera is mounted to the trolley platform moving between the two rows. In Scenario~2, the trolley vertically lifts and lowers the robot arm.}
	\label{fig:simulated_env}
\end{figure}

We design two glasshouse simulation scenarios resembling real-world deployment conditions as illustrated in \figref{fig:simulated_env}.
The simulation contains a robot arm mounted to a trolley platform moving horizontally along plant rows and vertically lifting the robot arm if needed. We simulate an RGB-D camera attached to the robot arm providing realistic point cloud observations.
The first simulation scenario consists of two plant rows on one vertical level with 12 plants and 47 fruits in total. The second scenario consists of two vertical levels with two plant rows and 12 plants each and 94 fruits in total. This scenario mimics deployment with tall sweet pepper plants. Each plant row is partitioned into four equidistant horizontal segments, 1\,m apart from each other. This way, Scenario~1 and Scenario~2 contain $8$ and $16$ segments, respectively. The trolley is positioned accordingly in each segment, and the arm's sampling and workspace are adjusted to each segment.

To evaluate our view motion planner's fruit monitoring performance in restricted workspaces, we compare it against our previously developed state-of-the-art next-best region of interest planner~\cite{zaenker2020viewpoint}. In contrast to our non-myopic graph-based path search, the RVP approach samples view pose candidates based on the current map state and greedily chooses the single next-best view pose using a utility function not accounting for the robot arm's motion constraints.
To assess the fruit monitoring performance under limited mission time, each planner is assigned $60$\,s to process a segment. We quantify the planning performance by counting detected fruits over the mission time including the planner's computation time and path execution time. As both planning approaches utilize stochastic view pose sampling, we conduct five experiment runs per planner in each scenario and report the mean and standard deviation of the performance metric.

\begin{figure}
	\centering
	\begin{subfigure}[b]{0.49\linewidth} 		\centering 		\includegraphics[width=\linewidth]{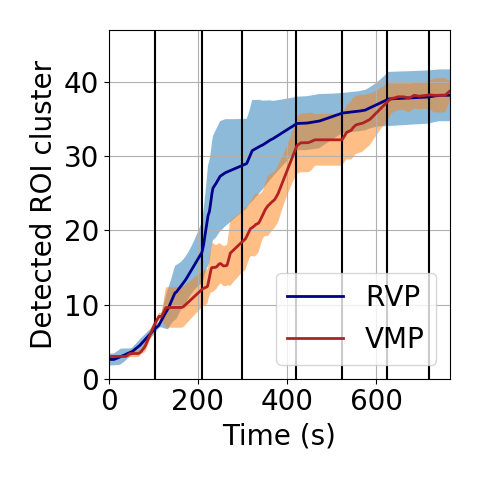} 		\caption{Scenario 1}
	\label{fig:res_sim_w22}
	\end{subfigure}
	\begin{subfigure}[b]{0.49\linewidth} 		\centering 		\includegraphics[width=\linewidth]{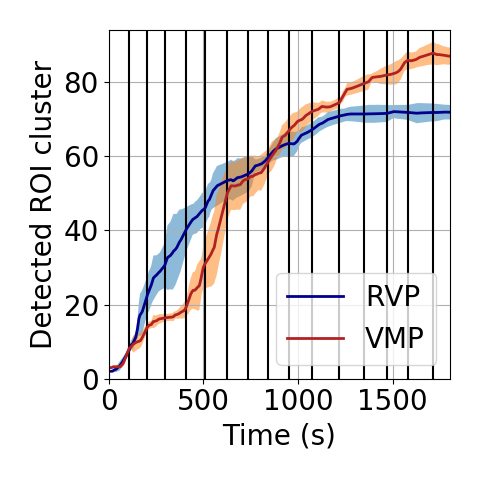} 		\caption{Scenario 2} 
	\label{fig:res_sim_w23}
	\end{subfigure}
	\caption{Simulated experimental results.
     RVP~(blue line) represents our old view pose planner which plans only the next-best view~\cite{zaenker2020viewpoint}.
    VMP~(red line) is our newly developed graph-based view motion planner.
    Black vertical lines mark segment changes after 60\,s exploration. Note that they are further apart since they include the transition to the next segment.} 
	\label{fig:res_sim}
\end{figure}


\subsection{Fruit Detection}
We aim to show the fruit detection performance of our VMP approach compared to the state-of-the-art RVP method after a fixed mission time. 
To evaluate the fruit detection performance of the two planners, we used our previously developed sweet pepper shape estimator~(Marangoz\etal\cite{marangoz2022fruit}).
The shape estimator runs Voxblox\cite{oleynikova2017voxblox} on the extracted fruit point clouds to obtain an accurate fruit map, then clusters the obtained surface point cloud and matches superellipsoids to estimate the fruit shapes and positions.
The estimated fruit positions are compared with the ground truth fruits extracted from the plant models of the simulation to verify that only correctly detected peppers are counted.

\figref{fig:res_sim} shows how the two planners vary in their approach to detecting fruit clusters over time.
In Scenario 1, VMP detects fruits at a steady rate, overcoming its deficit vis-a-vis RVP in the initial segments and thereby improving its performance towards the end of the mission, thanks to its systematic path search. 
RVP, on the other hand, finds distant view poses that cover parts of the neighboring plant segments as well, explaining the early advantage.
At the end of the mission, both planners achieve a similar fruit count ($38.2\pm3.9$ for RVP and $39.2\pm1.9$ for VMP out of the 47 detectable fruits).  
In Scenario 2, VMP outperforms RVP in detecting fruits. VMP detects on average $87.4\pm2.9$ 
compared to RVP's detection of $72.2\pm1.9$  
fruits out of a total of 94 fruits, in the end.
The performance of VMP in Scenario 2 is similar to that in Scenario 1, with a steady rate of fruit detection. Similar to Scenario 1, RVP detects greater number of clusters in the initial phase. However, it is not able to take advantage of its initial lead, and flattens out towards the end as it is unable to find sufficient view poses to detect new fruits within the limited mission time.

\subsection{View pose Execution Time}
In this section, we verify that our VMP reduces the time elapsed between two consecutive view poses crucial for monitoring efficiency. The measured time elapsed between two consecutive view poses includes both the planner's computation time and the motion execution time.

\begin{figure}[!t]
	\centering
	\includegraphics[width=0.95\linewidth]{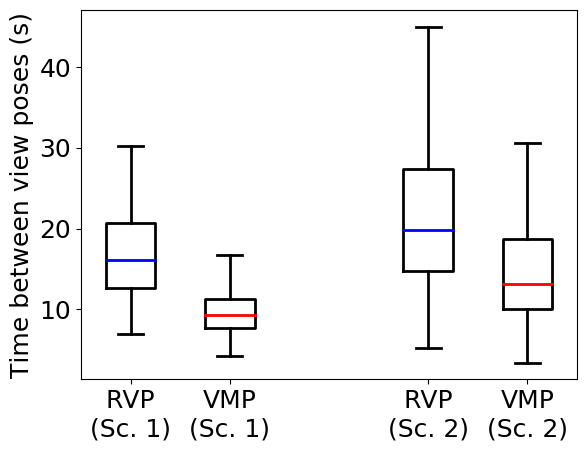}	
	\caption{Time between reaching new view poses.
 \textit{Left}: Comparison in Scenario 1 between the old (RVP) and our new planner (VMP).
 \textit{Right}: Scenario 2.
 In both cases, our new system finds new view poses in less time, which enables it to integrate more views within the given time budget to obtain a more complete map.}
	\label{fig:boxplot_vp_times}
\end{figure}

\figref{fig:boxplot_vp_times} shows the average time between reaching new view poses within the segments.
It is noticeable that VMP can execute more view poses within the same time budget.
It can be seen that VMP reaches new poses more quickly.
Especially in Scenario 2, where the workspace is more restricted, the number of outliers that need greater planning time increases for RVP leading to the execution of fewer informative view poses, as compared to VMP.
While the required time for VMP also increases, the outliers are fewer.
This could be due to the challenge of finding collision-free motion plans in such highly restricted environments.
Due to the myopic nature of next-best view planning, RVP greedily finds view poses that may have high information gain but not necessarily in the immediate vicinity of its current pose. This can lead to either repeated global motion planning failures, longer computation time, or motion execution time due to convoluted trajectories. All factors can lead to inefficient reachability of consecutive view poses.
VMP, on the other hand, connects only nearby view poses. 
Even if the next view pose needs to be discarded due to detection of a potential collision, the path lengths are smaller and the sequence is planned for a shorter horizon. 
This also explains why the advantage is especially obvious in Scenario 2:
due to the second vertical level, the environment is more constrained and VMP's limited sequence of nearby view poses helps it to shorten the time interval.
Also, since graph construction and path search can run in parallel, VMP can move to new view poses continuously while RVP replans at each pose.


\subsection{Real-World Glasshouse Experiments}
\label{sec:rwexp}

\begin{figure}
	\centering
	\includegraphics[width=\linewidth]{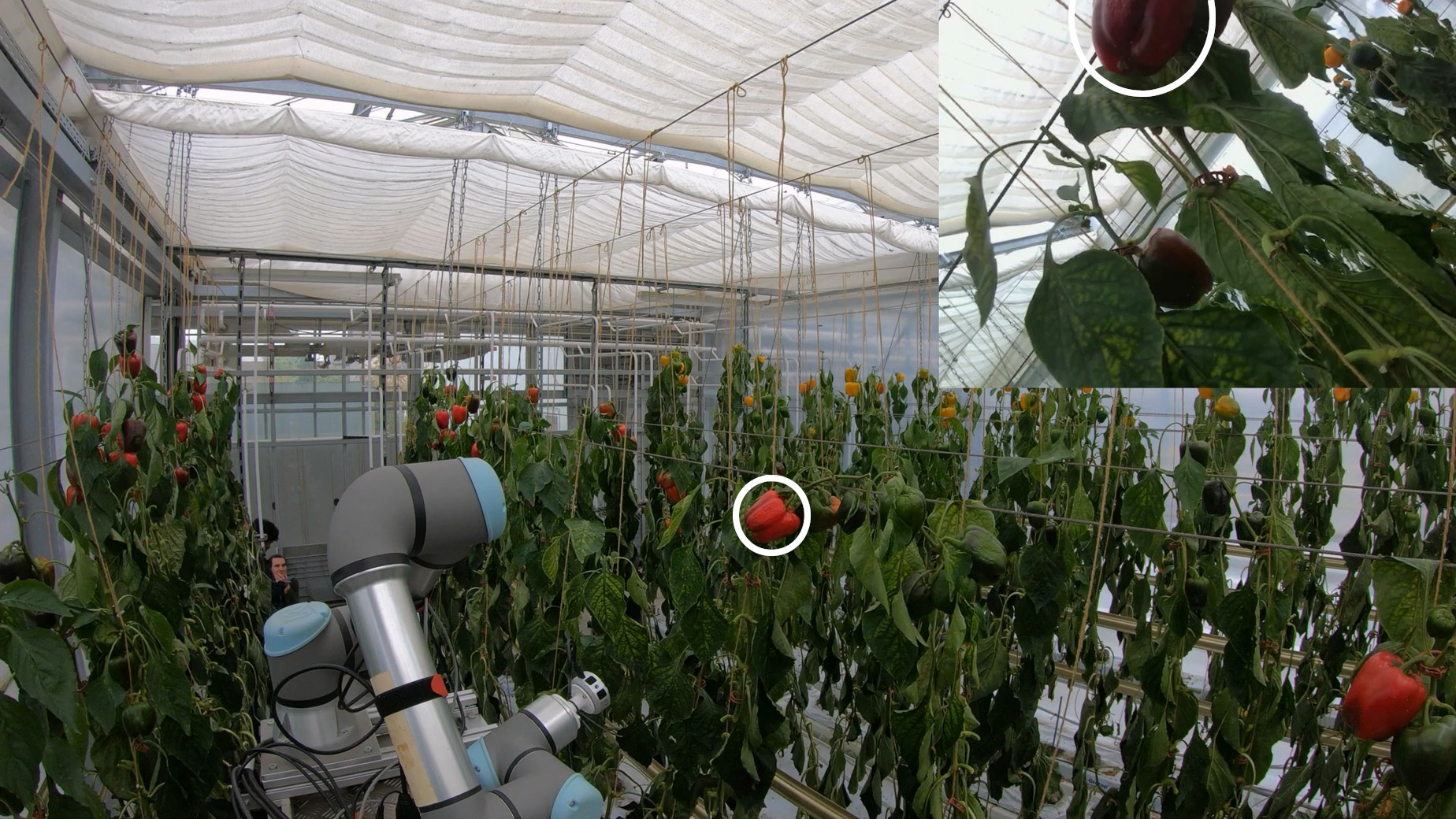}	
	\caption{Trolley in an indoor glasshouse.
	The main image shows the arm recorded with an external camera. In the top right, the view of the arm camera is shown.
    The white circles mark the same pepper in both views.
    The fruit that is hidden behind the leaf in the camera is not visible from the external view.}
	\label{fig:rwenv}
\end{figure}

We demonstrate the performance of our approach in a real-world indoor glasshouse mission mapping a sweet pepper plant row with a robot arm mounted on a trolley platform~\cite{mccool21icra}.
Details on the used hardware can be found in the referenced paper.

Similar to our simulation setup described in \cref{s:exp_setup}, we partitioned the row into equidistant segments monitoring the row segment by segment at a fixed platform height.
\figref{fig:rwenv} illustrates a planned view pose enabling the arm to observe a pepper fruit from below to avoid occlusions from the leaf in front of it verifying the benefit of our proposed view pose utility function.

\begin{figure}
	\centering
	\includegraphics[width=\linewidth]{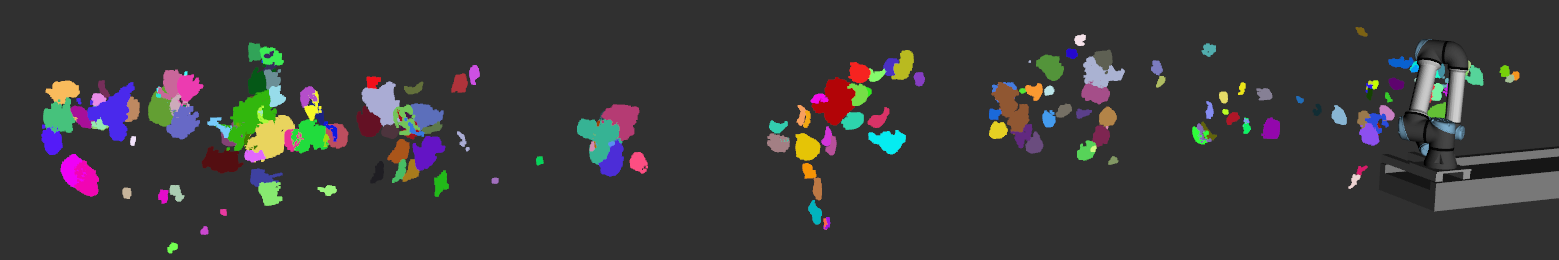}	
	\caption{Fruit clusters obtained using our view motion planner deployed on a platform in an indoor glasshouse. Our planning approach enables monitoring the majority of fruit clusters.}
	\label{fig:rw_fruit_clusters}
\end{figure}

\figref{fig:rw_fruit_clusters} shows the obtained fruit clusters along a plant row.
While no fruit ground truth is available, we qualitatively validate that our approach plans view poses monitoring the majority of fruit clusters.
Utilizing our clustering and fruit shape estimation approach~\cite{marangoz2022fruit}, the resulting fruit map could be leveraged in downstream tasks, e.g. yield estimation.

\begin{figure}
	\centering
	\sbox\twosubbox{
		\resizebox{\dimexpr.96\linewidth}{!}{
			\includegraphics[height=2.6cm]{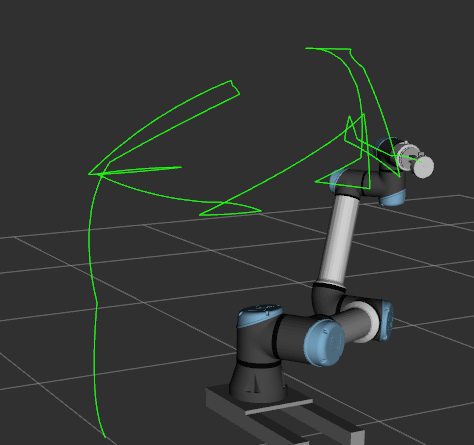}
			\includegraphics[height=2.6cm]{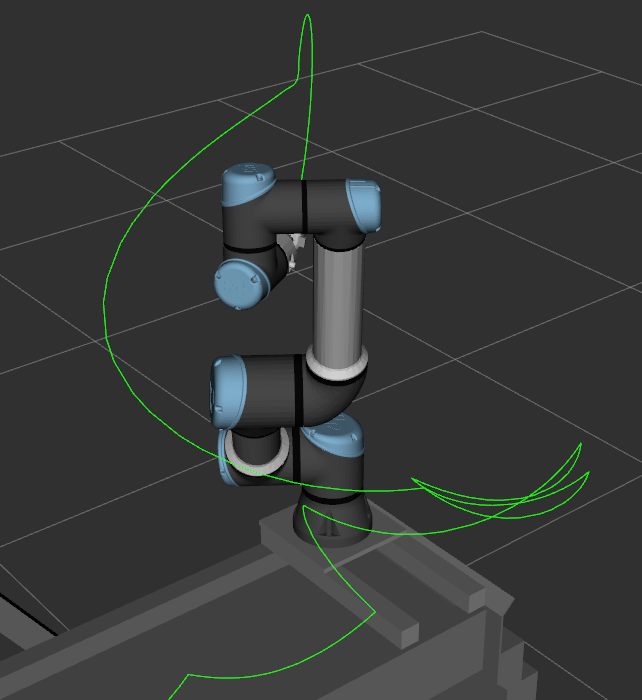}
		}
	}
	\setlength{\twosubht}{\ht\twosubbox}
	\centering
	\subcaptionbox{VMP path\label{fig:vmp_path_rw}}{
		\includegraphics[height=\twosubht]{images/vmp_path_rw}
	}\quad
	\subcaptionbox{RVP path\label{fig:rvp_path_rw}}{
		\includegraphics[height=\twosubht]{images/rvp_path_rw}
	}
	\caption{\textit{Left}: Example path obtained from the new view motion planner. \textit{Right}: Path resulting from our old next-best view planner~\cite{zaenker2020viewpoint}.
    In comparison, the new view pose sequence planner planned smaller segments that are more easily executed.}
	\label{fig:path_rw_comp}
\end{figure}

\figref{fig:path_rw_comp} showcases paths planned with our VMP approach compared to paths planned with the RVP method.
Our VMP approach generates shorter, straighter trajectories, whereas RVP plans more complicated and inefficient to execute paths to distant view poses. These results verify the general applicability of our approach in real-world glasshouse missions while previous methods fail to plan feasible paths or degrade in efficiency.

\section{Conclusions}
\label{sec:concl}

We presented a novel graph-based view motion planning approach for fruit monitoring using a robot arm in glasshouse environments.
Our planner constructs a graph of viable view pose candidates and connects neighboring, efficiently reachable poses.
It then non-myopically searches the graph for the view pose sequence that maximizes fruit cluster detections under consideration of limited motion capabilities in constrained workspaces.
The experimental results demonstrated that our new graph-based planner detects more fruit clusters in restricted robot arm workspaces given a limited time budget when compared against state-of-the-art single next-best view planning.
Moreover, we showed the real-world applicability of our approach for efficiently monitoring sweet peppers in an indoor glasshouse, where the small motion segments of our approach are essential for reliable execution.
In the future, we plan to integrate our system with more accurate mapping systems to improve the quality of the fruit reconstruction and enable a better estimation of the information gain along the view pose sequences.

\bibliographystyle{IEEEtran}
\bibliography{refs}

\end{document}